\documentclass[a4paper]{article}
\usepackage[top=30mm, bottom=30mm, left=20mm, right=20mm]{geometry}
\usepackage{indentfirst}
\usepackage{hyperref}
\usepackage{amsmath, amssymb, amsthm}
\usepackage{lipsum}
\usepackage[T1]{fontenc}
\usepackage{authblk}

\newtheorem*{mythm}{Theorem}
\newtheorem{myprop}{Proposition}
\newtheorem{apdxprop}{Proposition}
\newtheorem{mylem}{Lemma}
\newtheorem{apdxlem}{Lemma}
\DeclareMathOperator*{\argmin}{arg\,min}

\date{\today}

\title{Convergence Analysis of Optimization Algorithms}
\author[ ]{HyoungSeok Kim, JiHoon Kang, WooMyoung Park, SukHyun Ko \\ 
YoonHo Cho, DaeSung Yu, YoungSook Song and JungWon Choi}
\affil[ ]{Company.AI}
\affil[ ]{\texttt{\{alexrp, don, max.park, noag.go,\protect\\
ed.cho, dsds, song.cai, xiao\}\protect\\
@company.ai}}

\begin{document}

\maketitle

\begin{abstract}
The regret bound of an optimization algorithms is one of the basic criteria for evaluating the performance of the given algorithm. By inspecting the differences between the regret bounds of traditional algorithms and adaptive one, we provide a guide for choosing an optimizer with respect to the given data set and the loss function. For analysis, we assume that the loss function is convex and its gradient is Lipschitz continuous.
\end{abstract}

\section{Introduction}

Consider a minimizing problem of the convex objective function $J(\theta)$ with input parameter $\theta \in \Theta$, such as, 
\begin{equation}\label{min_J}
\min J(\theta)
\end{equation}
To get the minimizing argument of \eqref{min_J}, $\theta^* \in \Theta$, we use iterative methods to update the current parameter vector $\theta_{t}$. From the current step $t$, each method use the gradient of $J(\theta_t)$ with the step size $\eta$. Also, $\nabla_{\theta} J(\theta_{t})$ denotes the gradient of the objective function $J(\theta_t)$ at the current parameter at time step $t$ with respect to the parameter vector $\theta$. Generally, we define the loss function as $J(\theta) = f(\theta) + \varphi(\theta)$ where the convex instant loss $f(\theta)$ and the convex regularization function is $\varphi(\theta)$. For the analysis, we define the regret $R_J(T)$ as
\begin{equation}
	R_J(T) := \sum_{t=1}^T \Big[ J(\theta_t) - J(\theta^*) \Big]
\end{equation}
to estimate error bound. Also, to guarantee the convergence of the algorithms in this paper, we assume the convexity of $J$ and the $L$-Lipschitz continuous gradient of $J$ such as
\begin{itemize}
	\item $J$ is convex, i.e.
	\begin{equation}\label{J_convex}
		J(y) \geq J(x) + \langle \nabla J(x), y-x \rangle \quad \forall x, y
	\end{equation}
	\item $\nabla J(x)$ is $L$-Lipschitz continuous, i.e.
	\begin{equation}\label{J_lipschitz_cts}
		\| \nabla J(x) - \nabla J(y) \| \leq L \| x-y \|,\quad \forall x, y
	\end{equation}
	Also, the equation \eqref{J_lipschitz_cts} implies 
	\begin{equation}\label{J_lipschitz_grad}
		J(y) \leq J(x) + \langle \nabla J(x) , y-x \rangle + \frac{L}{2}\|y-x\|^2, \quad \forall x, y
	\end{equation}
\end{itemize}
The analysis mainly focuses on the regret bound of each algorithms. The choice of optimizer results the difference in the performance of the training procedure on the same neural network. Roughly, one can classify the optimization algorithm by its convergence rate. As the first order method, we have stochastic gradient descent(\autoref{sec:SGD}), momentum method(\autoref{sec:MMT}) and Nesterov accelerated gradient method(\autoref{sec:NAG}). For the adaptive method, Adagrad(\autoref{sec:ADG}), Adadelta, and Adam(\autoref{sec:ADM}) are well known. For the optimizing tasks such as training neural net, adaptive methods are usually preferred. But in recent research \cite[Figure 1]{Wilson} shows that the traditional first order algorithms such as stochastic gradient method or momentum method give better convergence results than the adaptive methods. One possible reason may lie on the structure of the estimating Hessian matrix in adaptive algorithms. This estimation issue will be mentioned later at \autoref{sec:ADG} briefly. 

\section{Stochastic Gradient Descent}\label{sec:SGD}

\subsection{The Updates}

The basic gradient descent optimization with a full batch is 
\begin{equation}\label{full_batch_GD}
	\theta_{t+1} = \theta_{t} - \eta \nabla_{\theta} J(\theta_{t})
\end{equation}
where $\eta$ is the learning rate. In contrast, stochastic gradient descent or mini-batch gradient descent algorithm updates the parameter vector for each data $i$ or $i^\textrm{th}$ mini-batch data set, such as
\begin{equation}\label{mini_batch_GD}
	\theta_{t+1} = \theta_{t} - \eta \nabla_{\theta} J(\theta_{t}; x_i, y_i)
\end{equation}
where $J(\theta_{t}; x_i, y_i)$ implies that we only have the partial information of our loss function. In other words, the partially given batch data guides the gradient direction for each iteration.

\subsection{Convergence Analysis}

In this section, we will show the regret bound of gradient descent algorithm with a full batch is bounded by some constant. Also, we will show that the stochastic gradient descent method shares the same regret bound. One can notice that the sequence $\{J(\theta_T)\}$ is not monotonically decreasing since our stochastic gradient does not guarantee the exact decreasing direction. Since we assume that the cost function $J$ is convex, a constant bound of $R_J(T)$ implies the error at a certain step is bounded by the inverse of the iteration number.

\begin{mythm}[Nestrov, 2.1.14]
	If $J(\theta)$ is convex and its gradient is $L$-Lipschitz continuous, then for $\eta \in (0, 1/L]$, the sequence $\{\theta_t\}$ generated by update \eqref{full_batch_GD} or \eqref{mini_batch_GD} satisfies 
	\begin{equation*}
		R_J(T) = O\Big(\|\theta_1 - \theta^*\|^2\Big)
	\end{equation*}
\end{mythm}

\begin{proof}
Since $J$ has $L$-Lipschitz continuity, by \eqref{J_lipschitz_grad}, we have 
	\begin{align*}
		J(\theta_{t+1}) &\leq J(\theta_{t}) + \langle \nabla_\theta J(\theta_{t}), \theta_{t+1} - \theta_{t} \rangle + \frac{L}{2} \| \theta_{t+1} - \theta_{t} \|^2 \\
		&= J(\theta_{t}) + \langle \nabla_\theta J(\theta_{t}), -\eta \nabla_\theta J(\theta_{t}) \rangle + \frac{L}{2} \| -\eta \nabla_\theta J(\theta_{t}) \|^2 \\
		&= J(\theta_{t}) - \eta \| \nabla_\theta J (\theta_{t}) \|^2 + \frac{\eta^2 L}{2} \| \nabla_\theta J (\theta_{t}) \|^2 \\
		&= J(\theta_{t}) - \eta \left( 1-\frac{\eta L}{2}\right) \| \nabla_\theta J (\theta_{t}) \|^2 \\
		&\leq J(\theta_{t}) - \frac{\eta}{2}\| \nabla_\theta J (\theta_{t}) \|^2 \qquad (\because \eta \in (0, 1/L]) \\
		&\leq J(\theta^*) + \langle \nabla_\theta J(\theta_{t}), \theta_{t} - \theta^* \rangle - \frac{\eta}{2}\|\nabla_\theta J(\theta_{t})\|^2 \qquad (\because J\textrm{ is convex}) \\
		&= J(\theta^*) + \langle \nabla_\theta J(\theta_{t}), \theta_{t} - \theta^* \rangle - \frac{\eta}{2}\|\nabla_\theta J(\theta_{t})\|^2 + \frac{1}{2\eta}\Big( \|\theta_t - \theta^* \|^2 - \|\theta_t - \theta^* \|^2 \Big) \\
		&= J(\theta^*) + \frac{1}{2\eta}\bigg( \|\theta_t - \theta^* \|^2 - \Big( \|\theta_t\|^2 - 2 \langle \theta_t, \theta^* \rangle + \|\theta^*\|^2 -2\eta \langle \nabla_\theta J (\theta_t) , \theta_t-\theta^* \rangle  + \eta^2 \|\nabla_\theta J(\theta_t)\|^2 \Big) \bigg) \\
		&= J(\theta^*) + \frac{1}{2\eta}\bigg( \|\theta_t - \theta^* \|^2 - \Big( \|\theta_t - \eta \nabla_\theta J(\theta_t)\|^2 - 2 \langle \theta_t-\eta \nabla_\theta J (\theta_t), \theta^* \rangle  + \|\theta^*\|^2 \Big) \bigg) \\
		&= J(\theta^*) + \frac{1}{2\eta}\Big( \|\theta_t - \theta^* \|^2 - \|\theta_{t+1} - \theta^*\|^2 \Big)
	\end{align*}
	Hence, we get
	\begin{equation}\label{regret}
		J(\theta_{t+1}) - J(\theta^*) \leq \frac{1}{2\eta}\Big( \|\theta_t - \theta^* \|^2 - \|\theta_{t+1} - \theta^*\|^2 \Big)
	\end{equation}
	Thus, apply \eqref{regret} to summing over the iterations,
	\begin{align*}
		\sum_{t=1}^T \Big[ J(\theta_t) - J(\theta^*) \Big] &\leq \frac{1}{2\eta}\sum_{t=1}^T \Big[ \|\theta_t - \theta^*\|^2 - \|\theta_{t+1} - \theta^*\|^2 \Big] \\
		&= \frac{1}{2\eta}\Big( \|\theta_1 - \theta^* \|^2 - \|\theta_{T+1} - \theta^* \|^2 \Big) \\
		&\leq \frac{1}{2\eta} \|\theta_1 - \theta^* \|^2
	\end{align*}
\end{proof}

\section{Momentum}\label{sec:MMT}

\subsection{The Updates}

To accelerate the convergence of gradient descent method, momentum method use the past steps to update the current step. Intuitively, the past steps are relevant to the next update and using this information seems natural. Here $\gamma$ is called momentum parameter and $\eta$ is the learning rate. The momentum update in \cite[(5)]{Yang} is as follows
\begin{equation}\label{momentum_update}
	\begin{split}
		v_{t+1} & = \gamma \ v_{t} - \eta \nabla_{\theta} J(\theta_{t}) \\
		\theta_{t+1} & = \theta_{t} + v_{t+1}
	\end{split}	
\end{equation}
In \cite[(4)]{Yang} The update equation \eqref{momentum_update} is equivalent to
\begin{equation}\label{momentum_one_line_update}
	\theta_{t+1} = \theta_t + \gamma (\theta_{t} - \theta_{t-1}) - \eta \nabla_\theta J(\theta_t) 
\end{equation}

\subsection{Convergence Analysis}

Since the momentum method modifies the basic structure of the gradient descent approach, they share the same convergence rate. Similar with the previous analysis, we assume $J(\theta)$ is convex and its gradient is $L$-Lipschitz continuous.

\begin{mythm}[Ghadimi, Theorem 1]
	If $J(\theta)$ is convex and its gradient is $L$-Lipschitz continuous, then for $\gamma \in [0, 1)$, $\eta\in (0, (1-\gamma)/L]$, the sequence $\{\theta_t\}$ generated by update \eqref{momentum_update} satisfies 
	\begin{equation*}
		R_J(T) = O\Big(\| \theta_{1} - \theta^* \|^2\Big)
	\end{equation*}
\end{mythm}

\begin{proof}
	For some $\gamma \in [0, 1)$, let
	\begin{equation*}
		p_t = \frac{\gamma}{1-\gamma}(\theta_t - \theta_{t-1})
	\end{equation*}
	where $t=1, 2, \cdots, T$ and assume that $\theta_0 = \theta_1$ and $p_0=0$. By \eqref{momentum_one_line_update},
	\begin{equation*}
		\theta_{t+1}+p_{t+1} = \frac{1}{1-\gamma}\theta_{t+1} - \frac{\gamma}{1-\gamma}\theta_t = \theta_t + p_t - \frac{\eta}{1-\gamma}\nabla_\theta J(\theta_t)
	\end{equation*}
	Consider the optimal solution as $\theta^*$. We have
	\begin{align*}
		\|\theta_{t+1} + p_{t+1} - \theta^* \|^2 &= \|\theta_{t} + p_{t} -\frac{\eta}{1-\gamma} \nabla_\theta J(\theta_t) - \theta^*\|^2 \\
		&= \|\theta_{t} + p_{t} - \theta^*\|^2 - \frac{2\eta}{1-\gamma}\langle \theta_{t} + p_{t} - \theta^*, \nabla_\theta J(\theta_t) \rangle + \left( \frac{\eta}{1-\gamma}\right)^2 \|\nabla_\theta J(\theta_t)\|^2 \\
		&= \|\theta_{t} + p_{t} - \theta^*\|^2 - \frac{2\eta}{1-\gamma} \langle \theta_t - \theta^*, \nabla_\theta J(\theta_t) \rangle \\
		&\quad - \frac{2\eta\gamma}{(1-\gamma)^2}\langle \theta_t - \theta_{t-1}, \nabla_\theta J(\theta_t) \rangle + \left( \frac{\eta}{1-\gamma}\right)^2 \|\nabla_\theta J(\theta_t)||^2
	\end{align*}
	Since $J(\theta)$ is convex function with $L$-Lipschitz continuous gradeint, we introduce following propositions from \cite[Theorem 2.1.5]{Nesterov}.
	\begin{myprop}[Nestrov, Theorem 2.1.5]\label{proposition_01}
		\begin{align*}
			0 \leq J(y) - J(x) - \langle \nabla J(x) , y-x \rangle &\leq \frac{L}{2} \| x-y \|^2 \\
			J(x) + \langle \nabla J(x), y-x \rangle + \frac{1}{2L}\| \nabla J(x) - \nabla J(y) \|^2 &\leq J(y)
		\end{align*}
	\end{myprop}
	The proof of above properties are provided in appendix \ref{proof_nesterov}. Substituting $x, y$, the above inequalities are modified as follows:
	\begin{align}
		J(\theta_{t}) - J(\theta_{t-1}) &\leq \langle \nabla_\theta J(\theta_{t}) , \theta_{t}-\theta_{t-1} \rangle \label{nesterov_thm1} \\
	J(\theta_t) -J(\theta^*) + \frac{1}{2L}\|\nabla_\theta J(\theta_t) \|^2 &\leq \langle \nabla_\theta J(\theta_t), \theta_t-\theta^* \rangle \label{nesterov_thm2} 
	\end{align}
	By \eqref{nesterov_thm2}, we obtain
	\begin{align*}
		\|\theta_{t+1} + p_{t+1} - \theta^* \|^2 &\leq \|\theta_{t} + p_{t} - \theta^*\|^2 - \frac{2\eta}{1-\gamma}\left( J(\theta_t) -J(\theta^*) + \frac{1}{2L}\|\nabla_\theta J(\theta_t) \|^2 \right) \\
		&\quad - \frac{2\eta\gamma}{(1-\gamma)^2}\langle \theta_t - \theta_{t-1}, \nabla_\theta J(\theta_t) \rangle + \left( \frac{\eta}{1-\gamma}\right)^2 \|\nabla_\theta J(\theta_t)\|^2
	\end{align*}
	Here, by \eqref{nesterov_thm1}, we get
	\begin{align*}
		\|\theta_{t+1} + p_{t+1} - \theta^* \|^2 &\leq \|\theta_{t} + p_{t} - \theta^*\|^2 - \frac{2\eta}{1-\gamma}\left( J(\theta_t) -J(\theta^*) + \frac{1}{2L}\|\nabla_\theta J(\theta_t) \|^2 \right) \\
		&\quad - \frac{2\eta\gamma}{(1-\gamma)^2} \Big(J(\theta_{t}) - J(\theta_{t-1})\Big) + \left( \frac{\eta}{1-\gamma}\right)^2 \|\nabla_\theta J(\theta_t)\|^2
	\end{align*}
	Adding $-\frac{2\eta\gamma}{(1-\gamma)^2}J(\theta^*)$ on both side and collecting the terms, we obtain
	\begin{align*}
		\left(\frac{2\eta}{1-\gamma} + \frac{2\eta\gamma}{(1-\gamma)^2}\right)\Big(J(\theta_t) - J(\theta^*)\Big) + \| \theta_{t+1} + p_{t+1} - \theta^* \|^2 &\leq \frac{2\eta\gamma}{(1-\gamma)^2}\Big(J(\theta_{t-1}) - J(\theta^*)\Big) + \| \theta_t + p_t - \theta^*\| \\
		&\quad +\frac{\eta}{1-\gamma}\left( \frac{\eta}{1-\gamma}-\frac{1}{L}\right)\|\nabla_\theta J(\theta_t) \|^2
	\end{align*}
	Since we assume $\eta\in (0, (1-\gamma)/L]$, the third term of right-hand-side is a negative value. Thus, the inequality should hold under the elimination of the third term. i.e.,
	\begin{equation*}
		\left(\frac{2\eta}{1-\gamma} + \frac{2\eta\gamma}{(1-\gamma)^2}\right)\Big(J(\theta_t) - J(\theta^*)\Big) + \| \theta_{t+1} + p_{t+1} - \theta^* \|^2 \leq \frac{2\eta\gamma}{(1-\gamma)^2}\Big(J(\theta_{t-1} - J(\theta^*)\Big) + \| \theta_t + p_t - \theta^*\|
	\end{equation*}
	Summing over $k=1, 2, \cdots, T$ gives
	\begin{align*}
		\frac{2\eta}{1-\gamma}\sum_{t=1}^T \Big[ J(\theta_t) - J(\theta^*) \Big] + \sum_{t=1}^T \left[ \frac{2\eta\gamma}{(1-\gamma)^2}\Big(J(\theta_t) - J(\theta^*)\Big) + \| \theta_{t+1} + p_{t+1} - \theta^* \|^2 \right] \\ 
		\leq \sum_{t=1}^T \left[ \frac{2\eta\gamma}{(1-\gamma)^2}\Big(J(\theta_{t-1}) - J(\theta^*)\Big) + \| \theta_{t} + p_{t} - \theta^* \|^2 \right]
	\end{align*}
	Since $\theta^*$ is the optimal solution of $J(\theta)$, every terms are positive, so that
	\begin{align*}
		\frac{2\eta}{1-\gamma}\sum_{t=1}^T \Big[ J(\theta_t) - J(\theta^*) \Big] \leq \frac{2\eta\gamma}{(1-\gamma)^2}\Big( J(\theta_{1}) - J(\theta^*) \Big) + \| \theta_{1} - \theta^* \|^2
	\end{align*}
\end{proof}
\section{Nesterov Accelerated Gradient}\label{sec:NAG}

\subsection{The Updates}

In \cite[(6)]{Yang} the standard update equations for NAG method is as follows:
\begin{equation}\label{nag_update}
	\begin{split}
		y_{t+1} & = \theta_{t} - \eta \nabla_{\theta} J(\theta_{t}) \\
		\theta_{t+1} &= y_{t+1} + \gamma (y_{t+1} - y_{t})
	\end{split}
\end{equation}
where $y_0 = \theta_0$. Here, we can understand NAG update more intuitively by modifying the same equation. By introducing $v_t = y_t - y_{t-1}$ with $y_0 = y_{-1}$, \eqref{nag_update} is equivalent to 
\begin{equation}\label{nag_update_2}
	\begin{split}
		v_{t+1} & = \gamma \ v_{t} - \eta \nabla_{\theta} J(y_{t} + \gamma \ v_{t}) \\
		y_{t+1} & = y_{t} + v_{t+1}
	\end{split}
\end{equation}
In \cite[(7)]{Yang}. Rather than updating $\theta_t$, in \eqref{nag_update_2} we update $y_t$ to minimize the objective function. The main idea for NAG is known as gamble first and correct later. As we can see in \eqref{nag_update_2}, NAG estimates the next point by jump through the previous gradient direction and calculates the gradient at that position to correct the estimated point.

\subsection{Convergence Analysis}

\begin{mythm}[Ghadimi, Theorem 3]
	If $J(\theta)$ is convex and its gradient is $L$-Lipschitz continuous, then for $\gamma \in [0, 1)$, $\eta\in (0, 1/L]$, the sequence $\{\theta_t\}$ generated by update \eqref{nag_update} satisfies 
	\begin{equation}
		R_J(T) = O\Big(\| \theta_{1} - \theta^* \|^2\Big)
	\end{equation}
\end{mythm}

\begin{proof}
	Let
	\begin{equation*}
		p_t = \frac{\gamma}{1-\gamma}\Big(\theta_t - \theta_{t-1} + \eta \nabla_\theta J(\theta_{t-1})\Big)
	\end{equation*}
	where $t=1, 2, \cdots, T$ and assume that $\theta_0 = \theta_1$ and $p_0 = 0$. By \eqref{nag_update}, yields
	\begin{align*}
		\theta_{t+1} + p_{t+1} = \frac{1}{1-\gamma}\theta_{t+1} + \frac{\gamma}{1-\gamma}\Big( \eta \nabla_\theta J(\theta_t) - \theta_t \Big) = \theta_t+p_t - \frac{\eta}{1-\gamma}\nabla_\theta J(\theta_t)
	\end{align*}
	Consider the optimal solution $\theta^*$, yields
	\begin{align*}
		\|\theta_{t+1} + p_{t+1} - \theta^*\|^2 &= \|\theta_{t} + p_{t} - \theta^*\|^2 
		- \frac{2\eta}{1-\gamma} \langle \theta_t + p_t - \theta^*, \nabla_\theta J (\theta_t) \rangle 
		+ \left(\frac{\eta}{1-\gamma} \right)^2 \| \nabla_\theta J (\theta_t) \|^2 \\
		&= \|\theta_{t} + p_{t} - \theta^*\|^2
		- \frac{2\eta}{1-\gamma} \langle \theta_t - \theta^*, \nabla_\theta J (\theta_t) \rangle
		- \frac{2\eta\gamma}{(1-\gamma)^2} \langle \theta_t - \theta_{t-1}, \nabla_\theta J (\theta_t) \rangle \\
		&\quad - \frac{2\eta^2\gamma}{(1-\gamma)^2} \langle \nabla_\theta J (\theta_{t-1}), \nabla_\theta J (\theta_t) \rangle
		+ \left(\frac{\eta}{1-\gamma} \right)^2 \| \nabla_\theta J (\theta_t) \|^2
	\end{align*}
	Again \cite[Theorem 2.1.5]{Nesterov}, we have \eqref{nesterov_thm2}. And also
	\begin{equation}\label{nesterov_thm3} 
	J(\theta_t) -J(\theta_{t-1}) + \frac{1}{2L}\|\nabla_\theta J(\theta_t)-\nabla_\theta J(\theta_{t-1}) \|^2 \leq \langle \nabla_\theta J(\theta_t), \theta_t-\theta_{t-1} \rangle 
	\end{equation}
	By \eqref{nesterov_thm2} and \eqref{nesterov_thm3}, yields
	\begin{align*}
		\|\theta_{t+1} + p_{t+1} - \theta^*\|^2 &\leq \|\theta_{t} + p_{t} - \theta^*\|^2 -\frac{2\eta}{1-\gamma} \left( J(\theta_t) - J(\theta^*) + \frac{1}{2L} \| \nabla_\theta J(\theta_t) \|^2 \right) \\
		&\quad - \frac{2\eta\gamma}{(1-\gamma)^2} \left( J(\theta_t) - J(\theta_{t-1}) + \frac{1}{2L} \| \nabla_\theta J(\theta_t)- \nabla_\theta J(\theta_{t-1}) \|^2 \right) \\
		&\quad - \frac{2\eta^2\gamma}{(1-\gamma)^2} \langle \nabla_\theta J (\theta_{t-1}), \nabla_\theta J (\theta_t) \rangle
		+ \left(\frac{\eta}{1-\gamma} \right)^2 \| \nabla_\theta J (\theta_t) \|^2
	\end{align*}
	Since $\eta\in (0, 1/L]$, we have
	\begin{align*}
		\|\theta_{t+1} + p_{t+1} - \theta^*\|^2 &\leq \|\theta_{t} + p_{t} - \theta^*\|^2 -\frac{2\eta}{1-\gamma} \left( J(\theta_t) - J(\theta^*) + \frac{\eta}{2} \| \nabla_\theta J(\theta_t) \|^2 \right) \\
		&\quad - \frac{2\eta\gamma}{(1-\gamma)^2} \left( J(\theta_t) - J(\theta_{t-1}) + \frac{\eta}{2} \| \nabla_\theta J(\theta_t)- \nabla_\theta J(\theta_{t-1}) \|^2 \right) \\
		&\quad - \frac{2\eta^2\gamma}{(1-\gamma)^2} \langle \nabla_\theta J (\theta_{t-1}), \nabla_\theta J (\theta_t) \rangle
		+ \left(\frac{\eta}{1-\gamma} \right)^2 \| \nabla_\theta J (\theta_t) \|^2
	\end{align*}
	Adding $\frac{2\eta\gamma}{(1-\gamma)^2} J(\theta^*)$ on both side, we get
	\begin{align*}
		\frac{2\eta}{(1-\gamma)^2}& \Big( J(\theta_t) - J(\theta^*) \Big) + \|\theta_{t+1} + p_{t+1} - \theta^*\|^2 \\ 
		&\leq \frac{2\eta\gamma}{(1-\gamma)^2} \Big( J(\theta_{t-1}) - J(\theta^*) \Big) + \|\theta_{t} + p_{t} - \theta^*\|^2 - \frac{\eta^2}{1-\gamma} \| \nabla_\theta J(\theta_t) \|^2 \\
		&\quad - \frac{\eta^2\gamma}{(1-\gamma)^2} \| \nabla_\theta J(\theta_t)- \nabla_\theta J(\theta_{t-1}) \|^2 - \frac{2\eta^2\gamma}{(1-\gamma)^2} \langle \nabla_\theta J (\theta_{t-1}), \nabla_\theta J (\theta_t) \rangle \\
		&\quad + \left(\frac{\eta}{1-\gamma} \right)^2 \| \nabla_\theta J (\theta_t) \|^2 \\
		&= \frac{2\eta\gamma}{(1-\gamma)^2} \Big( J(\theta_{t-1}) - J(\theta^*) \Big) + \|\theta_{t} + p_{t} - \theta^*\|^2 + \frac{\eta^2\gamma}{(1-\gamma)^2} \| \nabla_\theta J(\theta_t) \|^2 \\
		&\quad - \frac{\eta^2\gamma}{(1-\gamma)^2} \| \nabla_\theta J(\theta_t)- \nabla_\theta J(\theta_{t-1}) \|^2 - \frac{2\eta^2\gamma}{(1-\gamma)^2} \langle \nabla_\theta J (\theta_{t-1}), \nabla_\theta J (\theta_t) \rangle \\
		&= \frac{2\eta\gamma}{(1-\gamma)^2} \Big( J(\theta_{t-1}) - J(\theta^*) \Big) + \|\theta_{t} + p_{t} - \theta^*\|^2 \\
		&\quad + \frac{\eta^2\gamma}{(1-\gamma)^2} \Big( \| \nabla_\theta J(\theta_t) \|^2 - \| \nabla_\theta J(\theta_t)- \nabla_\theta J(\theta_{t-1}) \|^2 - 2 \langle \nabla_\theta J (\theta_{t-1}), \nabla_\theta J (\theta_t) \rangle \Big) \\
		&= \frac{2\eta\gamma}{(1-\gamma)^2} \Big( J(\theta_{t-1}) - J(\theta^*) \Big) + \|\theta_{t} + p_{t} - \theta^*\|^2 \\
		&\quad + \frac{\eta^2\gamma}{(1-\gamma)^2} \Big( \| \nabla_\theta J(\theta_t) \|^2 - \| \nabla_\theta J(\theta_t)- \nabla_\theta J(\theta_{t-1}) \|^2 - 2 \langle \nabla_\theta J (\theta_{t-1}), \nabla_\theta J (\theta_t) \rangle \\
		&\quad + \| \nabla_\theta J(\theta_{t-1}) \|^2 - \| \nabla_\theta J(\theta_{t-1}) \|^2 \Big) \\
		&= \frac{2\eta\gamma}{(1-\gamma)^2} \Big( J(\theta_{t-1}) - J(\theta^*) \Big) + \|\theta_{t} + p_{t} - \theta^*\|^2 \\
		&\quad + \frac{\eta^2\gamma}{(1-\gamma)^2} \Big(  \| \nabla_\theta J(\theta_t)- \nabla_\theta J(\theta_{t-1}) \|^2 - \| \nabla_\theta J(\theta_t)- \nabla_\theta J(\theta_{t-1}) \|^2  - \| \nabla_\theta J(\theta_{t-1}) \|^2 \Big) \\
		&= \frac{2\eta\gamma}{(1-\gamma)^2} \Big( J(\theta_{t-1}) - J(\theta^*) \Big) + \|\theta_{t} + p_{t} - \theta^*\|^2 - \frac{\eta^2\gamma}{(1-\gamma)^2}\| \nabla_\theta J(\theta_{t-1}) \|^2
	\end{align*}
	Multiplying $1/2\eta$ on both side and summing over $t = 1, 2, \cdots, T$ gives
	\begin{align*}
		\frac{1}{1-\gamma} \sum_{t=1}^T \Big[ J(\theta_t) - J(\theta^*) \Big] + \sum_{t=1}^T &\left[ \frac{\gamma}{(1-\gamma)^2} \Big( J(\theta_t) - J(\theta^*) \Big) + \frac{1}{2\eta} \|\theta_{t+1} + p_{t+1} - \theta^*\|^2 \right] \\
		&\leq \sum_{t=1}^T \left[ \frac{\gamma}{(1-\gamma)^2} \Big( J(\theta_{t-1}) - J(\theta^*) \Big) + \frac{1}{2\eta} \|\theta_{t} + p_{t} - \theta^*\|^2 \right]
	\end{align*}
	Therefore we have
	\begin{equation*}
		\frac{1}{1-\gamma} \sum_{t=1}^T \Big[ J(\theta_t) - J(\theta^*) \Big] \leq \frac{\gamma}{(1-\gamma)^2} \Big( J(\theta_{0}) - J(\theta^*) \Big) + \frac{1}{2\eta} \|\theta_{1} - \theta^*\|^2
	\end{equation*}
	where $\theta_0 = \theta_1$ by assumption.
\end{proof}

\section{Adagrad}\label{sec:ADG}

Including Adagrad method, the adaptive method in the next sections follow the Newton's method which is known as the second-order method. Since these methods minimize the objective function $J$ with estimated Hessian matrix and apply the Newton's method approach, they generally perform better than above algorithms. Usually the cost of exact calculation of the Hessian matrix is extremely expensive, therefore Adagrad algorithm estimates the Hessian matrix with the following idea. According to the \cite[5.4.2]{Bishop}, consider the mean squared error function, such as
\begin{align*}
	J = \frac{1}{2} \sum_{n=1}^N (f(\theta_n) - y_n)^2
\end{align*}
Thus, the gradient and Hessian of $J$ is
\begin{align*}
	\nabla J &= \sum_{n=1}^N \langle f(\theta_n) - y_n, \nabla f(\theta_n) \rangle \\
	H(J) &= \sum_{n=1}^N \langle \nabla f(\theta_n), \nabla f(\theta_n) \rangle +  \sum_{n=1}^N \langle f(\theta_n) - y_n, \Delta f(\theta_n) \rangle
\end{align*}
Here the second term of Hessian equation goes to zero when the approximation of $f(\theta_n)$ close to the real value $y_n$, which implies estimate the Hessian matrix with the outer product of the gradient vector. This approximation is quite reasonable under the given mean squared error functions. But this approximation does not always proper under the arbitrary designed cost functions. Especially for the classification tasks, we often use the non-smooth cost functions such as Cross Entropy loss. Consequently, as we mention in the introduction, this estimation causes potential limitation of adaptive methods that applied in various loss functions.

Additionally, one of the benefit in Adagrad which the author of \cite{Duchi} mentioned is since the method updates the parameter vector element-wisely, Adagrad can perform better than previous methods like SGD or momentum method when the loss function $J$ is sparse. Compare with dense cases, sparse $J$ has relatively more chance to get the sparse gradient vector. And with the Adagrad method, that gives the larger step size, so that the gradient direction highly affects to the optimization process. Therefore, rarely occurring factor has more importance than frequently occurring factors.

\subsection{The Updates}
We use second sub-script for the vector or matrix element index. i,e, $\theta_{t, i}$ means the $i$th parameter of parameter vector at time step $t$. 

Since the convexity of $J$ does not imply the differentiability of $J$, we import the concept of sub-gradient. The sub-gradient can be applied to all of algorithms covered here. The sub-differentiable set of function $J$ evaluated at $\theta$ is denoted as $\partial J(\theta)$, and a particular gradient vector in the sub-gradient set is denoted by $g_t \in \partial J(\theta)$. When a function $J$ is differentiable, $g_{t}$ directly implies $\nabla_{\theta} J(\theta_{t})$. We also denote $g_{1:t} = [ g_1, g_2, \cdots, g_t]$ the concatenated matrix of the subgradient sequence.

The important feature in Adagrad is calculate the outer product of sub-gradient, denoted by $G_t \in \mathbb{R}^{d \times d}$ where $d$ is the number of entry in $\theta$, means
\begin{equation}
	G_{t} = \sum_{\tau = 1}^{t} g_{\tau}^{\phantom{\intercal}}g_{\tau}^{\intercal}
\end{equation}
As we mentioned before, Adagrad method element-wisely updates parameter vector. In \cite[(1)]{Duchi}, Adagrad update the parameter such as
\begin{equation}\label{Adagrad}
	\theta_{t+1, i} = \theta_{t, i} - \frac{\eta} {\sqrt{G_{t, ii} + \epsilon}} \cdot g_{t, i}
\end{equation}

\subsection{Convergence Analysis}

For the analysis, we convert the form of update equation \eqref{Adagrad}. By \cite[(1)]{Duchi}, consider a Euclidean space $\Theta$ and convert the update equation \eqref{Adagrad} as
\begin{equation}\label{adagrad_update_mid}
	\theta_{t+1} = \argmin_{\theta \in \Theta} \big\| \theta - (\theta_t - \eta \ \textrm{diag}(G_t)^{-1/2} g_t )\big\|^2_{G_t^{1/2}}
\end{equation}
where the Mahalanobis norm $\| \cdot \|_A = \sqrt{\langle \cdot, A \cdot \rangle}$ and $A^{1/2}$ implies the element-wise root of given matrix or vector $A$. Next, introduce the Bregman divergence associated with a strongly convex function $\psi$, which is
\begin{equation}
	B_\psi (x, y) = \psi(x) - \psi(y) - \langle \nabla \psi (y), x-y \rangle
\end{equation}
According to the \cite[(3), (4)]{Duchi}, claim that for some regularization function $\varphi$, we can convert \eqref{adagrad_update_mid} as
\begin{equation}\label{adagrad_update_final}
	\theta_{t+1} = \argmin_{\theta \in \Theta} \Big\{ \eta \langle g_t, \theta \rangle + \eta \varphi(\theta) + B_{\psi_t}(\theta, \theta_t) \Big\}
\end{equation}
to update our parameter vector $\theta$.

\begin{mythm}[Duchi, Theorem 5] If $J(\theta)$ is convex and its gradient is $L$-Lipschitz continuous, then for $\theta^* \in \Theta$, the sequence $\{\theta_t\}$ which generated by \eqref{adagrad_update_final} satisfies
	\begin{equation*}
		R_J(T) = O\Big( \max_{t\leq T} \|\theta_t-\theta^*\|_\infty \sum_{t=1}^d \|g_{1:T, i} \|_2 \Big)
	\end{equation*}
\end{mythm}

\begin{proof} Let $g_t$ be defined as in above. We have the following proposition and the proof is in \cite[Appendix F]{Duchi}.
	\begin{myprop}[Duchi, Proposition 3]\label{proposition_02}
		Let the sequence $\{\theta_t\}$ be defined by the update \eqref{adagrad_update_final}. For any $\theta^* \in \Theta$,
		\begin{equation*}
			R_J(t) \leq \frac{1}{\eta} B_{\psi_t} ( \theta^*, \theta_1) + \frac{1}{\eta}\sum_{t=1}^{T-1} \Big[ B_{\psi_{t+1}}(\theta^*, \theta_{t+1}) - B_{\psi_t}(\theta^*, \theta_{t+1}) \Big]+\frac{\eta}{2}\sum_{t=1}^T \|J'(\theta_t)\|^2_{\psi_t^*}
		\end{equation*}
	\end{myprop}
	
	Let $s_t$ is a vector at time step $t$ such that $i$th element of the vector $s_{t, i} = \|g_{1:t, i} \|_2$. The following lemma is proved in appendix \ref{proof_duchi_lem}.
	\begin{mylem}[Duchi, Lemma 4]\label{lemma_01}
		Let $g_t$, $g_{1:t}$ and $s_t$ be defined as in above. Then
		\begin{equation*}
			\sum_{t=1}^T \big\langle g_t, \mathrm{diag} (s_t)^{-1} g_t \big\rangle \leq 2 \sum_{i=1}^d \|g_{1:T, i}\|_2
		\end{equation*}
	\end{mylem}	
	Here, define the associated dual-norm of $\psi_t(x)$
	\begin{equation*}
		\|g\|^2_{\psi_t^*} = \big\langle g, (\delta I + \mathrm{diag} (s_t))^{-1} g \big\rangle
	\end{equation*}
	where $\psi_t(x)= \langle x, (\delta I + \mathrm{diag}(s_t))x\rangle$. Since $g_t$ is a subgradient of $J(\theta)$, implies $\|J'(\theta_t)\| \leq \langle g_t, \textrm{diag}(s_t)^{-1} g_t \rangle$. Thus, yield,
	\begin{equation*}
		\sum_{t=1}^T \|J'_t(\theta_t)\|^2_{\psi_t^*} \leq 2 \sum_{i=1}^d \|g_{1:T, i}\|_2
	\end{equation*}
	Now, the Bregman divergence terms in above proposition are remained. We notice that
	\begin{align*}
		B_{\psi_{t+1}}(\theta^*, \theta_{t+1}) - B_{\theta_t}(\theta^*, \theta_{t+1}) &= \frac{1}{2}\big\langle \theta^*-\theta_{t+1}, \textrm{diag}(s_{t+1}-s_t)(\theta^*-\theta_{t+1} )\big\rangle \\
		&\leq \frac{1}{2}\max_i (\theta^*_i -\theta_{t+1,i})^2 \| s_{t+1} - s_t\|_1
	\end{align*}
	Since $\| s_{t+1} - s_{t} \|_1 = \langle s_{t+1} - s_t, 1\rangle$ and $\langle s_T, 1 \rangle = \sum_{i=1}^d \| g_{1:T, i} \|_2$, we have
	\begin{align*}
		\sum_{t=1}^{T-1} \Big[ B_{\psi_{t+1}} (\theta^*, \theta_{t+1}) - B_{\psi_{t}} (\theta^*, \theta_{t+1}) \Big]	&\leq \frac{1}{2}\sum_{t=1}^{T-1} \|\theta^* - \theta_{t+1}\|_{\infty}^2 \langle s_{t+1}- s_t, 1 \rangle \\
		&\leq \frac{1}{2}\max_{t \leq T} \| \theta^* - \theta_t \|_{\infty}^2 \sum_{i=1}^d \| g_{1:T,i} \|_2 - \frac{1}{2} \| \theta^* - \theta_1 \|^2_{\infty} \langle s_1, 1 \rangle
	\end{align*}
	Combine the proposition and using the above results with the fact that $B_{\psi_1}(\theta^*, \theta_1) \leq \frac{1}{2} \|\theta^* - \theta_1\|_\infty^2\langle s_1, 1 \rangle$, we finally get
	\begin{equation*}
		R_J(T) \leq \frac{1}{2\eta} \max_{t \leq T} \| \theta^* - \theta_t \|_\infty^2 \sum_{i=1}^d \| g_{1:T,i} \|_2 + \eta \sum_{i=1}^d \| g_{1:T,i} \|_2
	\end{equation*}
\end{proof}

\section{Adam}\label{sec:ADM}

\subsection{The Updates}
Consider the estimates of the first and the second moment of the gradients. In \cite[Algorithm 1]{Kingma}, for some $\beta_1, \beta_2 \in [0, 1)$,
\begin{equation}\label{adam_momentum}
	\begin{split}
		m_t & = \beta_1 m_{t-1} + (1 - \beta_1)g_t \\
		v_t & = \beta_2 v_{t-1} + (1 - \beta_2)g_t^2
	\end{split}
\end{equation}
The authors of this method said $m_t$ and $v_t$ are biased towards zero especially during the initial stages and when the decay rates are small (i.e. $\beta_1$ and $\beta_2$ are nearly 1). So we need bias-correction, such as
\begin{equation}\label{adam_bais}
	\begin{split}
		\hat{m}_t & = \frac{m_t}{1-\beta_1^t} \\
		\hat{v}_t & = \frac{v_t}{1-\beta_2^t} \\
	\end{split}
\end{equation}
The final update equation is
\begin{equation}\label{adam_update}
	\theta_{t+1} = \theta_{t} - \frac{\eta}{\sqrt{\hat{v}_t} + \epsilon} \hat{m}_t
\end{equation}

\subsection{Convergence Analysis}
We show the regret bound of Adam method with learning rate $\eta_t$ is decaying at a rate of $\sqrt{t}$ and moment average coefficient $\beta_1$ decays exponentially with $\lambda$. 
\begin{mythm}[Kingma, Theorem 4.1]
	If $J(\theta)$ is convex and its gradient is $L$-Lipschitz continuous, i.e., $\| \nabla J(\theta) \|_2 \leq L$, $\| \nabla J(\theta) \|_\infty \leq L_\infty$ for all $\theta \in \Theta$ and for any $m, n \in \{1, 2, \cdots, T\}$, $\| \theta_m - \theta_n \|_2 \leq D$, $\| \theta_m - \theta_n \|_\infty \leq D_\infty$ then for all $T \geq 1$, the sequence $\{\theta_t\}$ which generated by \eqref{adam_momentum}, \eqref{adam_bais}, and \eqref{adam_update} satisfies  
	\begin{equation*}
		R_J(T) = O(\sqrt{T})
	\end{equation*}
	where $\beta_1, \beta_2 \in [0, 1)$ satisfy $\beta_1^2 / \sqrt{\beta_2} < 1$ and $\eta_t = \eta/\sqrt{t}, \eta_0 = \eta$ in the update equations.
\end{mythm}

\begin{proof} The following lemmas are used to support the theorem above. The proofs are in appendix \ref{proof_kingma_lem}
	\begin{mylem}[Kingma, lemma 10.4]\label{kingma_lemma}
		Let $\gamma := \beta_1^2 / \sqrt{\beta_2}$. For $\beta_1, \beta_2 \in [0, 1)$ that satisfy $\gamma < 1$ and bounded $g_t$, i.e., $\|g_t\|_2 \leq L$, $\|g_t\|_\infty \leq L_\infty$, the following holds
		\begin{equation*}
			\sum_{t=1}^T \frac{\hat{m}^2_{t, i}}{\sqrt{t\hat{v}_{t, i}}} \leq \frac{2L_\infty}{(1-\gamma)^2\sqrt{1-\beta_2}}\|g_{1:T, i}\|_2
		\end{equation*}
		where $\hat{m}_t$ and $\hat{v}_t$ are defined in \ref{adam_bais}
	\end{mylem}	
	Since our cost function $J$ is convex, we have
	\begin{equation*}
		J(\theta_t) - J(\theta^*) \leq \langle g_t, \theta_t - \theta^* \rangle = \sum_{i=1}^d g_{t, i}\cdot(\theta_{t, i} - \theta_i^*)
	\end{equation*}
	From the update rules, for some $\lambda\in(0, 1)$
	\begin{align*}
		\theta_{t+1} &= \theta_t - \eta_t \frac{\hat{m}_t}{\sqrt{\hat{v}_t}} \\
		&= \theta_t - \frac{\eta_t}{1-\beta_1^t}\left( \frac{\beta_1 \lambda^{t-1}}{\sqrt{\hat{v}_t}} m_{t-1} + \frac{1- \beta_1 \lambda^{t-1}}{\sqrt{\hat{v}_t}} g_{t} \right)
	\end{align*}
	Now, consider $i$th element of $\theta_t$ in Euclidean vector space. On both side of the update equation, we subtract $\theta^*_i$ and square, yield
	\begin{align*}
		(\theta_{t+1, i} - \theta^*_i)^2 &= (\theta_{t, i} - \theta^*_i)^2 - 2\eta_t\frac{\hat{m}_t}{\sqrt{\hat{v}_t}}(\theta_{t, i} - \theta^*_i) + \eta_t^2 \left(\frac{\hat{m}_{t, i}}{\sqrt{\hat{v}_{t, i}}}\right)^2 \\
		&= (\theta_{t, i} - \theta^*_i)^2 - \frac{2\eta_t}{1-\beta_1^t}\left( \frac{\beta_1 \lambda^{t-1}}{\sqrt{\hat{v}_{t, i}}} m_{t-1, i} + \frac{1-\beta_1 \lambda^{t-1}}{\sqrt{\hat{v}_{t, i}}}g_{t, i}\right) (\theta_{t, i} - \theta^*_i) + \eta_t^2 \left(\frac{\hat{m}_{t, i}}{\sqrt{\hat{v}_{t, i}}}\right)^2
	\end{align*}
	Using the fact that $2ab \leq a^2 + b^2$, yield
	\begin{align*}
		g_{t, i} \cdot (\theta_{t, i} - \theta^*_i) 
		&= \frac{(1-\beta_1^t) \sqrt{\hat{v}_{t, i}}}{2\eta_t(1-\beta_1 \lambda^{t-1})}\Big( (\theta_{t, i} - \theta^*_i)^2 - (\theta_{t+1, i} - \theta^*_i)^2\Big) \\
		&\quad + \frac{\beta_1\lambda^{t-1}}{1-\beta_1\lambda^{t-1}}(\theta^*_{i} - \theta_{t, i})m_{t-1, i} + \frac{\eta_t (1-\beta_1^t)(\hat{m}_{t, i})^2}{2(1-\beta_1\lambda^{t-1})\sqrt{\hat{v}_{t, i}}}\\
		&= \frac{(1-\beta_1^t) \sqrt{\hat{v}_{t, i}}}{2\eta_t(1-\beta_1 \lambda^{t-1})}\Big( (\theta_{t, i} - \theta^*_i)^2 - (\theta_{t+1, i} - \theta^*_i)^2\Big) \\
		&\quad + \frac{\beta_1\lambda^{t-1}}{1-\beta_1\lambda^{t-1}}(\theta^*_{i} - \theta_{t, i})\frac{\sqrt[4]{\hat{v}_{t-1, i}}}{\sqrt{\eta_{t-1}}}\frac{\sqrt{\eta_{t-1}}}{\sqrt[4]{\hat{v}_{t-1, i}}}m_{t-1, i} + \frac{\eta_t (1-\beta_1^t)}{2(1-\beta_1\lambda^{t-1})}\frac{(\hat{m}_{t, i})^2}{\sqrt{\hat{v}_{t, i}}} \\
		&\leq \frac{\sqrt{\hat{v}_{t, i}}}{2\eta_t(1-\beta_1)}\Big( (\theta_{t, i} - \theta^*_i)^2 - (\theta_{t+1, i} - \theta^*_i)^2\Big) \\
		&\quad + \frac{\beta_1\lambda^{t-1}}{1-\beta_1\lambda^{t-1}}(\theta^*_{i} - \theta_{t, i})^2 \frac{\sqrt{\hat{v}_{t-1, i}}}{2\eta_{t+1}} + \left( \frac{\beta_1}{1-\beta_1} \right)\frac{\eta_{t+1}(\hat{m}_{t-1, i})^2}{2\sqrt{\hat{v}_{t-1, i}}} + \frac{\eta_t}{2(1-\beta_1)}\frac{(\hat{m}_{t, i})^2}{\sqrt{\hat{v}_{t, i}}}
	\end{align*}
	One can notice that $\hat{v}_{t, i} = \sum_{\tau=1}^t (1-\beta_2)\beta_2^{t-\tau}g^2_{\tau,i} / (1-\beta_2^t) \leq \|g_{1:t, i} \|^2_2$. Intuitively, the exponentially decaying weighted sum must be less than or equal to the general summation of a given sequence. We apply the lemma \ref{kingma_lemma} to the above inequality and derive the regret bound by summing over all the dimensions for $i = 1, 2, \cdots, d$ in $J(\theta_t) - J(\theta^*)$ and the sequence of regrets for $t = 1, 2, \cdots, T$. The index of the summation in following inequality is modifying the above inequality by adding or subtracting the initial or the final term of some sequences to match the index unity.
	\begin{align*}
		R_J(T) &\leq \sum_{t=1}^T \sum_{i=1}^d g_{t, i}\cdot(\theta_{t, i} - \theta_i^*) \\
		&\leq \frac{1}{2\eta_t(1-\beta_1)}\sum_{i=1}^d (\theta_{1, i} - \theta^*_i)^2 \sqrt{\hat{v}_{1, i}} + \frac{1}{2\eta_t(1-\beta_1)}\sum_{i=1}^d \sum_{t=2}^T (\theta_{t, i}  - \theta^*_i)^2(\sqrt{\hat{v}_{t, i}} - \sqrt{\hat{v}_{t-1, i}}) \\
		&\quad +\sum_{i=1}^d \sum_{t=1}^T \frac{\beta_1\lambda^{t-1}}{2\eta_t(1-\beta_1\lambda^{t-1})}(\theta_{t, i} - \theta^*_i)^2 \sqrt{\hat{v}_{t, i}} + \frac{\beta_1 \eta L_\infty}{(1-\beta_1)\sqrt{1-\beta_2}(1-\gamma)^2} \sum_{i=1}^d \|g_{1:T, i}\|^2 \\
		&\quad + \frac{\eta L_\infty}{(1-\beta)\sqrt{1-\beta_2}(1-r)^2} \sum_{i=1}^d \|g_{1:T, i}\|^2
	\end{align*}
	From the assumption, $\|\theta_t - \theta^* \|_2 \leq D, \|\theta_m - \theta_n \|_\infty \leq D_\infty$. Also 
	\begin{align*}
		R_J(T) &\leq \frac{D^2}{2\eta(1-\beta_1)} \sum_{i=1}^d \sqrt{T \hat{v}_{T, i}} +\frac{(D_\infty)^2 L_\infty}{2\eta}\sum_{i=1}^d \sum_{t=1}^T \frac{\beta_1\lambda^{t-1}}{1-\beta_1\lambda^{t-1}} \sqrt{t} + \frac{\eta (\beta_1 + 1) L_\infty}{(1-\beta_1)\sqrt{1-\beta_2}(1-\gamma)^2} \sum_{i=1}^d \|g_{1:T, i}\|^2
	\end{align*}
	The upper bound of the arithmetic geometric series yields
	\begin{align*}
		\sum_{t=1}^T \frac{\beta_1\lambda^{t-1}}{1-\beta_1\lambda^{t-1}} \sqrt{t} &\leq \sum_{t=1}^T\frac{1}{1-\beta_1}\lambda^{t-1}\sqrt{t} \\
		&\leq \frac{1}{1-\beta_1}\sum_{t=1}^T \lambda^{t-1} t = \frac{1 + \lambda + \lambda^2 + \cdots + \lambda^{T-1} - \lambda^TT}{(1-\beta_1)(1-\lambda)}\\
		&\leq \frac{1 + \lambda + \lambda^2 + \cdots }{(1-\beta_1)(1-\lambda)} = \frac{1}{(1-\beta_1)(1-\lambda)^2}
	\end{align*}		
	Therefore, we have the following regret bound as
	\begin{align*}
		R_J(T) &\leq \frac{D^2}{2\eta(1-\beta_1)} \sum_{i=1}^d \sqrt{T \hat{v}_{T, i}} + \sum_{i=1}^d \frac{(D_\infty)^2 L_\infty}{2\eta(1-\beta_1)(1-\lambda)^2} + \frac{\eta (\beta_1 + 1) L_\infty}{(1-\beta_1)\sqrt{1-\beta_2}(1-\gamma)^2} \sum_{i=1}^d \|g_{1:T, i}\|^2
	\end{align*}
\end{proof}

\section*{Appendix}
\appendix
\section{Proof of Proposition \ref{proposition_01}}
\label{proof_nesterov} 
\begin{apdxprop}
	For the convex $J(\theta)$ with $L$-Lipschitz continuous gradient, implies the following inequalities
	\begin{align}
		0 \leq J(y) - J(x) - \langle \nabla J(x) , y-x \rangle &\leq \frac{L}{2} \| x-y \|^2 \label{prop_1_1}\\
		J(x) + \langle \nabla J(x), y-x) + \frac{1}{2L}\| \nabla J(x) - \nabla J(y) \|^2 &\leq J(y)\label{prop_1_2}
	\end{align}
\end{apdxprop}	
	
\begin{proof}
	Clearly, \ref{prop_1_1} comes from the definition of the convex function and $L$-Lipschitz continuous gradient. The remaining part is \ref{prop_1_2}. For some fixed $x_0 \in \Theta$, consider a function $g(x) = J(x) - \langle \nabla J(x_0), x \rangle$. Then
	\begin{align*}
		g(y) - g(x) - \langle \nabla g(z), y-x \rangle &= J(y) - \langle \nabla J(x_0), y \rangle - J(x) + \langle \nabla J(x_0), x \rangle - \langle \nabla J(x) - \nabla J(x_0), y-x \rangle \\
		&= J(y) - J(x) - \langle \nabla J(x) , y-x \rangle
	\end{align*}
	Thus, $g(x)$ is also convex function with $L$-Lipschitz gradient and its optimal point $x^*=x_0$. Therefore, applying second inequality of \ref{prop_1_1} to $g(x)$, yields
	\begin{align*}
		g\left(y-\frac{1}{L}\nabla g(y) \right) - g(y) -\left\langle \nabla g(y), y-\frac{1}{L}\nabla g(y) -y \right\rangle \leq \frac{1}{2L}\|\nabla g(y) \|^2
	\end{align*}
	Since $x^*$ is an optimal point of $g(x)$, we have
	\begin{align*}
		g(x_0)=g(x^*) \leq g\left(y-\frac{1}{L}\nabla g(y) \right) \leq g(y) - \frac{1}{2L}\|\nabla g(y) \|^2
	\end{align*}
	From $\nabla g(y) = \nabla J(y) - \nabla J(x_0)$, we get
	\begin{align*}
		J(x_0) - \langle \nabla J(x_0), x_0 \rangle \leq J(y) - \langle \nabla J(x_0), y \rangle - \frac{1}{2L}\| \nabla J(y) - \nabla J(x_0) \|^2
	\end{align*}
	Since we start with arbitrary $x_0$ as a dummy variable, we finally get the inequality
	\begin{align*}
		J(x) + \langle \nabla J(x), y-x \rangle + \frac{1}{2L}\| \nabla J(x) - \nabla J(y) \|^2 \leq J(y)
	\end{align*}
\end{proof}

\section{Proof of Lemma \ref{lemma_01}}
\label{proof_duchi_lem}
\begin{apdxlem}
	Let $g_t$, $g_{1:t}$ and $s_t$ be defined above. Then
	\begin{equation*}
		\sum_{t=1}^T \big\langle g_t, \mathrm{diag} (s_t)^{-1} g_t \big\rangle \leq 2 \sum_{i=1}^d \|g_{1:T, i}\|_2
	\end{equation*}
\end{apdxlem}	
	
\begin{proof}
	We prove the lemma by considering an arbitrary real-valued sequence $\{ a_i \}$ and $a_{1:t} = [a_1, a_2, \cdots, a_i]$. Consider,
	\begin{align*}
		\sum_{t=1}^T \frac{(a_t)^2}{\| a_{1:t} \|_2} \leq 2 \| a_{1:T} \|_2
	\end{align*}
	We use an induction on $T$ to prove the above inequality. For $T=1$, the inequality is clear. Assume the inequality holds for $T-1$, by the induction assumption,
	\begin{align*}
		\sum_{t=1}^T \frac{(a_t)^2}{\| a_{1:t} \|_2} = \sum_{t=1}^{T-1} \frac{(a_t)^2}{\| a_{1:t} \|_2} + \frac{(a_T)^2}{\| a_{1:T} \|_2} \leq 2 \| a_{1:T} \|_2 + \frac{(a_T)^2}{\| a_{1:T} \|_2}
	\end{align*}
	Suppose $b_T = \sum_{t=1}^T (a_t)^2$ and we obtain
	\begin{align*}
		\sqrt{b_T - (a_T)^2} \leq \sqrt{b_T - (a_T)^2 + \frac{(a_T)^4}{4b_T}} = \sqrt{b_t} - \frac{(a_T)^2}{2\sqrt{b_T}}
	\end{align*}
	Thus, we have
	\begin{align*}
		2\|a_{1:T-1}\|_2  + \frac{(a_T)^2}{\| a_{1:T} \|_2} = 2\sqrt{b_T - (a_T)^2} + \frac{(a_T)^2}{\sqrt{b_T}} = 2\|a_{1:T}\|_2
	\end{align*}
	Note that by construction that $s_{t,i} = \|g_{1:t, i}\|_2$. so
	\begin{align*}
		\sum_{t=1}^T \langle g_t, \textrm{diag}(s_t)^{-1}g_t \rangle = \sum_{t=1}^T \sum_{i=1}^d \frac{(g_{t, i})^2}{\|g_{1:t, i}\|_2} \leq 2\sum_{i=1}^d \| g_{1:T, i} \|_2
	\end{align*}
\end{proof}

\section{Proof of Lemma \ref{kingma_lemma}}
\label{proof_kingma_lem}
Before we begin the proof, we will prove the following lemma first.
\begin{mylem}[Kingma, Lemma 10.3]\label{lemma_03_apdx}
	Let $g_t = \nabla J(\theta_t)$ and $g_{1:t} = [ g_1, g_2, \cdots, g_t]$ is bounded. i.e., $\|g_t\|_2 \leq L$, $\|g_t\|_\infty \leq L_\infty$. Then
	\begin{equation*}
		\sum_{t=1}^T \sqrt{\frac{g^2_{t, i}}{t}} \leq 2 L_\infty \|g_{1:T, i}\|_2
	\end{equation*}
\end{mylem}

\begin{proof}
	We will prove the inequality using induction over T. For $T=1$, we have
	\begin{align*}
		g_{1,i} \leq 2 G_\infty \|g_{1, i}\|_2
	\end{align*}
	For the induction, we assume the following is true.
	\begin{align*}
		\sum_{t=1}^{T-1} \sqrt{\frac{g^2_{t, i}}{t}} \leq 2 L_\infty \|g_{1:T-1, i}\|_2
	\end{align*}
	Then we have,
	\begin{align*}
		\sum_{t=1}^{T} \sqrt{\frac{g^2_{t, i}}{t}} 
		&\leq 2 L_\infty \|g_{1:T-1, i}\|_2 + \sqrt{\frac{g^2_{T, i}}{T}} \\
		&= 2 L_\infty \sqrt{\|g_{1:T, i}\|_2^2 - (g_{T,i})^2} + \sqrt{\frac{g^2_{T, i}}{T}}
	\end{align*}
	We want to show the last equation is less than $ 2 L_\infty \|g_{1:T, i}\|_2$. From the fact that 
	\begin{align*}
		\|g_{1:T, i}\|_2^2 - (g_{T,i})^2 \leq \|g_{1:T, i}\|_2^2 - (g_{T,i})^2 + \frac{(g_{T, i})^4}{4\| g_{1:T, i} \|_2^2}
	\end{align*}		
	We take the square root at both side. Since $\| g_t \|_2 \leq \| g_t \|_2 \leq L_\infty$, we have
	\begin{align*}
			\sqrt{\|g_{1:T, i}\|_2^2 - (g_{T,i})^2} 
			&\leq \|g_{1:T, i}\|_2 - \frac{(g_{T, i})^2}{2\|g_{1:T, i \|_2}} \\
			&\leq \|g_{1:T, i}\|_2 - \frac{(g_{T, i})^2}{2\sqrt{T( L_\infty)^2}}
	\end{align*}
	Therefore, substituting the root term, yields
	\begin{align*}
		2 L_\infty \sqrt{\|g_{1:T, i}\|_2^2 - (g_{T,i})^2} + \sqrt{\frac{g^2_{T, i}}{T}} \leq 2L_\infty \|g_{1:T, i} \|_2
	\end{align*}
\end{proof}

\begin{apdxlem}\label{lemma_02_apdx}
	Let $\gamma := \beta_1^2 / \sqrt{\beta_2}$. For $\beta_1, \beta_2 \in [0, 1)$ that satisfy $\gamma < 1$ and bounded $g_t$, i.e., $\|g_t\|_2 \leq L$, $\|g_t\|_\infty \leq L_\infty$, the following holds
	\begin{equation*}
		\sum_{t=1}^T \frac{\hat{m}^2_{t, i}}{\sqrt{t\hat{v}_{t, i}}} \leq \frac{2L_\infty}{(1-\gamma)^2\sqrt{1-\beta_2}}\|g_{1:T, i}\|2
	\end{equation*}
	where $\hat{m}_t$ and $\hat{v}_t$ are defined in \ref{adam_bais}
\end{apdxlem}		

\begin{proof} 
	Under the assumption $\frac{\sqrt{1-\beta_2^t}}{(1-\beta_1)^2} \leq \frac{1}{(1-\beta_1)^2}$. We can expand the last term.
	\begin{align*}
		\sum_{t=1}^T \frac{(\hat{m}_{t,i})^2}{\sqrt{t\hat{v}_{t,i}}} 
		&= \sum_{t=1}^{T-1} \frac{(\hat{m}_{t,i})^2}{\sqrt{t\hat{v}_{t,i}}} + \frac{\sqrt{1-\beta_2^T}}{(1-\beta_1^T)^2} \frac{\left(\sum_{k=1}^T (1-\beta_1)\beta_1^{T-k} g_{k, i}\right)^2}{\sqrt{T\sum_{j=1}^T (1-\beta_2) \beta_2^{T-j} (g_{j,i})^2}} \\
		&\leq \sum_{t=1}^{T-1} \frac{(\hat{m}_{t,i})^2}{\sqrt{t\hat{v}_{t,i}}} + \frac{\sqrt{1-\beta_2^T}}{(1-\beta_1^T)^2} \sum_{k=1}^T \frac{T \left( (1-\beta_1)\beta_1^{T-k} g_{k, i}\right)^2}{\sqrt{T\sum_{j=1}^T (1-\beta_2) \beta_2^{T-j} (g_{j,i})^2}} \\
		&\leq \sum_{t=1}^{T-1} \frac{(\hat{m}_{t,i})^2}{\sqrt{t\hat{v}_{t,i}}} + \frac{\sqrt{1-\beta_2^T}}{(1-\beta_1^T)^2} \sum_{k=1}^T \frac{T \left( (1-\beta_1)\beta_1^{T-k} g_{k, i}\right)^2}{\sqrt{T (1-\beta_2) \beta_2^{T-k} (g_{k,i})^2}} \\
		&\leq \sum_{t=1}^{T-1} \frac{(\hat{m}_{t,i})^2}{\sqrt{t\hat{v}_{t,i}}} + \frac{\sqrt{1-\beta_2^T}}{(1-\beta_1^T)^2} \frac{T(1-\beta_1)^2}{\sqrt{T(1-\beta_2)}}\sum_{k=1}^T \left( \frac{\beta_1^2}{\sqrt{\beta_2}} \right)^{T-k} \|g_{k,i} \|_2 \\
		&\leq \sum_{t=1}^{T-1} \frac{(\hat{m}_{t,i})^2}{\sqrt{t\hat{v}_{t,i}}} + \frac{T}{\sqrt{T(1-\beta_2)}}\sum_{k=1}^T \gamma^{T-k} \|g_{k,i} \|_2 
	\end{align*}
	Expanding the rest of the terms in summation, yields
	\begin{align*}
		\sum_{t=1}^T \frac{(\hat{m}_{t,i})^2}{\sqrt{t\hat{v}_{t,i}}}  
		&\leq \sum_{t=1}^T \frac{\|g_{t, i}\|_2}{\sqrt{t(1-\beta_2)}}\sum_{j=0}^{T-t} t\gamma^j \\
		&\leq \sum_{t=1}^T \frac{\|g_{t, i}\|_2}{\sqrt{t(1-\beta_2)}}\sum_{j=0}^{T} t\gamma^j		
	\end{align*}
	For $\gamma < 1$, the sum of arithmetic-geometric series is bounded as $\sum_t t\gamma^t \leq 1/(1-\gamma)^2$, which yields
	\begin{align*}
		\sum_{t=1}^T \frac{\|g_{t, i}\|_2}{\sqrt{t(1-\beta_2)}}\sum_{j=0}^{T} t\gamma^j \leq \frac{1}{(1-\gamma)^2\sqrt{1-\beta_2}}\sum_{t=1}^T \frac{\|g_t, i\|_2}{\sqrt{t}}
	\end{align*}
	Finally, we apply the lemma \ref{lemma_03_apdx}, yielding
	\begin{align*}
		\sum_{t=1}^T \frac{(\hat{m}_{t,i})^2}{\sqrt{t\hat{v}_{t,i}}} \leq \frac{2L_\infty}{(1-\gamma)^2\sqrt{1-\beta_2}}\|g_{1:T,i} \|_2
	\end{align*}
\end{proof}

\end{document}